# Adversarial Attacks and Defences for Skin Cancer Classification

Joy Purohit
Department Of Computer Engineering
and Information Technology
Veermata Jijabai Technological Institute
Mumbai, India
purohitjoy21@gmail.com

Ishaan Shivhare
Department Of Computer Engineering
and Information Technology
Veermata Jijabai Technological Institute
Mumbai, India
ishaanshivhare2001@gmail.com

Vinay Jogani
Department Of Computer Engineering
and Information Technology
Veermata Jijabai Technological Institute
Mumbai, India
joganivinay@gmail.com

Samina Attari
Department Of Computer Engineering
and Information Technology
Veermata Jijabai Technological Institute
Mumbai, India
sameenaattari7860@gmail.com

Shraddha Surtkar
Department Of Computer Engineering
and Information Technology
Veermata Jijabai Technological Institute
Mumbai, India
sssuratkar@ce.vjti.ac.in

*Abstract*—There has been a concurrent significant improvement in the medical images used to facilitate diagnosis and the performance of machine learning techniques to perform tasks such as classification, detection, and segmentation in recent years. As a result, a rapid increase in the usage of such systems can be observed in the healthcare industry, for instance in the form of medical image classification systems, where these models have achieved diagnostic parity with human physicians. One such application where this can be observed is in computer vision tasks such as the classification of skin lesions in dermatoscopic images. However, as stakeholders in the healthcare industry, such as insurance companies, continue to invest extensively in machine learning infrastructure, it becomes increasingly important to understand the vulnerabilities in such systems. Due to the highly critical nature of the tasks being carried out by these machine learning models, it is necessary to analyze techniques that could be used to take advantage of these vulnerabilities and methods to defend against them. This paper explores common adversarial attack techniques. The Fast Sign Gradient Method (FGSM) and Projected Descent Gradient (PGD) are used against a Convolutional Neural Network (CNN) trained to classify dermatoscopic images of skin lesions. Following that, it also discusses one of the most popular adversarial defense techniques, adversarial training. The performance of the model that has been trained on adversarial examples is then tested against the previously mentioned attacks, and recommendations to improve neural network's robustness are thus provided based on the results of the experiment.

I. Introduction

The recent years have seen a substantial increase in the usage of machine learning techniques for a vast number of valuable applications. There has been a greater level of integration of such processes in the workflows of important industries in a way that has never been seen before. One such vital application is in the context of healthcare, where highly accurate machine learning models have found a role in the health information economy. For instance, diagnostic parity with human physicians has been achieved by artificial intelligence systems on various medical tasks in the domains of ophthalmology, radiology, and dermatology [1]. Furthermore, the first-ever artificial intelligence (AI) medical diagnosis system received marketing approval, thus highlighting the extent of the applications such systems could see and are currently seeing in the healthcare domain. As the usage of such systems becomes more ubiquitous in critical processes, it is important to study the vulnerabilities of such systems as well as the techniques that can be used to defend against their misuse.

The name given to individuals or entities that could make an effort to compromise or breach a computer network or application is "adversary". An "adversarial attack" is an attack that aims to compromise a machine learning model. This may be done either in the training phase or after the model has already been trained. This paper focuses on the second case, which would be the preferred method of choice for a majority of actors attempting to defraud machine learning systems in the medical context. These kinds of attacks are carried out using adversarial examples.

Adversarial examples are examples which are deliberately created to force the machine learning model to make an erroneous prediction. This is carried out by taking real data and engineering it in such a way that the intentional changes to the data will fool the processing algorithm. In cases of text-based data, this is done by adding text that may look quite innocent or substituting certain words with their synonyms in such a way that it could cause a misclassification. In other cases, such as those dealing with image-based data, this is carried out by adding tiny small perturbations to the data pixel-wise that are challenging to notice with the naked eye.

Machine learning has a direct impact on the healthcare model that can be observed in relation to the approval of insurance claims. Due to the enormous volume of medical claims that are processed every year, and the value in processing these medical claims, insurance firms and their contractors have made significant investments in infrastructure related to machine learning. Given that there are multiple competing financial interests vested in the outcomes of such claims, it is not unreasonable to anticipate that certain actors would engage in fraudulent practices to exploit vulnerabilities in such systems.

One instance of such a machine learning system is a medical image classification system. As more and more insurance companies have begun to require additional data to validate claims, such as images, there is a need to process these applications with a high degree of efficiency. Extremely accurate medical image classification systems that use advanced techniques for computer vision such as the deep learning method Convolutional Neural Network (CNN), fulfill these requirements. However, many types of machine learning models, such as neural networks, are extremely susceptible to attacks that are based on a slight modification to the input of the model during testing. Gradient-based techniques can be used to cause misclassifications in these algorithms. As demonstrated in the paper [2], neural networks find extreme difficulty in resisting linear adversarial perturbation due to their linear nature. The technique used by the authors of the aforementioned paper, known as Fast Sign Gradient Method (FGSM), along with another method known as Projected Gradient Descent (PGD), is among the techniques that can be used to create adversarial examples. Such attacks can be defended against. [2] explains how to make neural networks can be made more resistant to adversarial attacks via robust optimization, and it offers the initial steps toward completely resilient deep learning based models.

In this paper, the analysis of FGSM and PGD as techniques for adversarial attacks is carried out on a CNN model trained to classify skin lesions. The model is trained on the Skin Cancer MNIST: HAM1000 Dataset [3]. Following that, adversarial training is implemented as a defense technique. In essence, dataset augmentation is performed on the data used for the purpose of the training model for this paper using the adversarial techniques mentioned previously, and analysis is carried out on the results obtained in the conducted experiments.

## II. RELATED WORK

[2] proposes that, due to the extreme nature of their linearity, it is possible to describe adversarial examples as a property of high-dimensional dot products. The assertion that adversarial perturbations are strongly aligned with the weight vectors supports the generalization of adversarial examples in various models. The direction of perturbation is more important than specific points in space. The paper also talks about how adversarial perturbations generalize because the direction is so important. A fast method that can be create adversarial examples is proposed called the FGSM attack method.

[4] deals with the problem of the weakness of deep learning models which comes with adversarial attacks. The paper proposes adversarial robustness through the use of robust optimization. The proposed methodology helps to identify methods for defense and attack that are reliable. It also gives security against any adversary. This also helps train significantly improved resistance against a wide range of such attacks. Further, the paper proposes the project Gradient method which is a first-order adversary attack.

[5] the paper proposes a fast adversarial training algorithm that is a robust model with almost no extra cost coming with natural training. The key changes to the original algorithm are to update both the model parameters and image perturbations with the help of simultaneous backward passes in place of gradient computations for each step. The proposed algorithm has the same computational cost as the conventional training but it is faster than previously stated methods.

Since the existing methods can only fool black box models, the paper [6] proposes a novel momentum-based algorithm to boost the attacks. The use of the momentum-based proposed algorithm will stabilize the update directions to counter the problem of poor local maxima iterations. The algorithm is also applied to an ensemble model to improve the success rate and also to prove that adversely trained models with strong defense are vulnerable to the proposed attacks.

## III. MEDICAL IMAGE CLASSIFICATION

Medical imaging has had a tremendous impact on the field of medicine, as these images grant the ability to perform a diagnosis of diseases that may be afflicting the subjects in need. With regards to the task of classification of these images, Deep Neural Networks (DNN) have emerged as an extremely efficient method in machine learning to improve diagnosis using these images and assist healthcare professionals. Examples include performing nodule segmentation on CT images and categorizing diabetic retinopathy from optical coherence tomography (OCT) [11, 12]. These medical DNNs have achieved performance parity with humans and their diagnoses are the same as those of human staff [10]. For the purpose of this paper, we will be performing our experiments on a CNN model trained to classify skin lesions using dermatoscopic images of the same.

### A. Model

We have employed CNN as our model. The model has been trained on the HAM100000 dataset to diagnose skin lesions from their dermatoscopic images as one of 7 different cases. CNN classification has been used in various medical classification systems and is a popular model of choice for such tasks. Convolutional Neural Networks are a variant of Deep Neural Networks [14].

For classification, our model employs a fully connected layer, dropout layers, batch normalization layers, max-pooling layers, and five convolution layers. A cross-entropy loss function is then used to direct the model and modify the weights before the classification is completed using the softmax activation function as the last layer.

### B. Dataset

The "Humans Against Machines with 10000 Training Images" (HAM10000) [3] dataset was used to train and test our model. The dataset contains 10015 dermatoscopic images in total. Our training data consists of 8012 images, and testing data consists of 2003 samples. There are seven classes of examples in the dataset, and are representative of all important categories. The size of each image has been changed to 28x28x3. All of the images are RGB images, and the pixel values have been scaled between 0 and 1. A class imbalance has been observed in this dataset. So, in order to fix this, random oversampling was applied to the training images, increasing their total size to 37359.

## IV. ADVERSARIAL ATTACK

Inputs that have been deliberately constructed to fool a machine learning model are called adversarial examples. For example, these examples force it to make a misclassification. The attack works on the principle of adding noise to an input image, such that the model's prediction will be modified from the intended class to a different class. These attacks can cause problems in the healthcare domain where medical image classification systems are in use, and are particularly insidious due to the fact that these perturbations are invisible to the human eye [7]. This means that a genuine input image and an adversarial example are indistinguishable when looked at by a human.

According to Ren, K et. al. [9] two categories that we can divide adversarial attacks into:

● Black-box Attack: Here, an adversary attempts to undermine the targeted model without any access to its parameters. Instead, the adversary relies on a trial-and-error-based approach and makes use of random assumptions or calculated guesses.

● White-box Attack: In this instance, the opponent has access to the parameters of the underlying model. The attack can significantly impact the model's performance because the adversary is aware of the model's training policy.

For the purpose of our paper, we are exploring a particular class of white-box attacks known as Gradient-based adversarial attacks. These attacks take advantage of a simple concept that is fundamental to back-propagation, which is an algorithm that deep neural networks are trained using.

The determination of the error between the network's output and the desired output, with regard to a particular input, is the first stage of the back-propagation process. Now, gradients are calculated corresponding to each parameter by using the error that was calculated in the previous step, keeping the input constant. These gradients are then utilized in order to change and update the parameters at each step of the process of training, to minimize the loss.

Gradient-based adversarial attacks exploit this process by making a slight change to the algorithm, thus generating a perturbation vector for the input image. Instead of considering the input to be constant, the attack considers the input to be variable and the gradients to be constant, hence obtaining gradients corresponding to each element of the input. These can be used to generate the perturbation vector [13]. In order to maximize the loss, the parameters are then updated in the direction of these gradients. Hence, by superimposing a factor of the generated noise onto the original input image, we are able to create an adversarial example specific to the target model.

The adversarial attacks explored in our paper have been discussed and explained below.

### A. Fast Gradient Sign Method

The Fast Gradient Sign Method (FGSM), first forward in [2], is one of the first, simplest, and fastest adversarial assault tactics. This technique, which is an adversarial approach based on gradients, operates by computing the gradients across the image and the loss between the target label and predicted label. As seen in equation 1, the gradient's "sign" is used to create noise.

$$x\_adv = x + \epsilon * sign(\nabla_x L(\theta, x, y))$$
(1)

Equation (1) depicts the mathematical equation of FGSM, where 'x' and 'y' represent images and labels for the same. As the attack updates the weight of the model in the direction of the gradient to result in a misclassification, the variable "sign," which represents the perturbation, is essentially the sign of the gradient with respect to "x." '$\epsilon$' denotes the size of the perturbation. It is a small value that is multiplied by the perturbation in order to keep the attacked image nearly impossible to differentiate from the original to the human eye, and L is the loss function in use by the model is denoted by L.

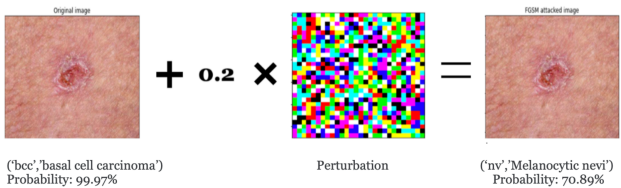

Figure 1: FGSM attack

As illustrated in figure 1, on the left-hand side we observe a dermatoscopic image 'x' where the lesion is classified by the model to be of the class 'basal cell carcinoma' with a probability of 99.97%. The perturbation 'sign' is then generated using the FGSM attack, after which it is multiplied with '$\epsilon$' and added to the original image that is denoted by 'x'.

The attacked image 'x_adv' is now classified as 'melanocytic nevi' with a probability of 70.89%. The attack successfully fools the model into misclassifying the image with a confidence of 70.89%.

### B. Project Gradient Descent

The Projected Gradient Descent [4] method follows the same logic as FGSM, but removes the restriction of a single-step update. PGD maximizes the model loss, with the restriction that the perturbation '$\delta$' is smaller than the value specified for '$\epsilon$'.

This is denoted by the mathematical expression that can be seen below:

$$\delta = \max_{\|\delta\|_p \leq \epsilon} (L(X + \delta, Y))$$
(2)

PGD enables choosing the best possible $\epsilon$-bounded perturbation. This $\epsilon$ is to be chosen such that the perturbation is indistinguishable. The PGD algorithm consists of the following steps:

1. Start from anywhere within Lp Ball around the sample, X
2. Take a gradient step in the direction of the highest gradient of Loss(X, Y)
3. Project steps back into Lp Ball
4. Repeat steps 2,3 till convergence

Every time an update occurs, the step is reflected back into the Lp ball surrounding the sample. Lp denotes the space where all points are within $\epsilon$ distance from the sample. The projection of a point, z, into the Lp ball is simply finding the nearest point to z in the Lp ball. Since the choice of p is not critical for a finite-dimensional space (such as most input spaces), the equivalence of norms ensures that tuning epsilon has the same effect as varying p.

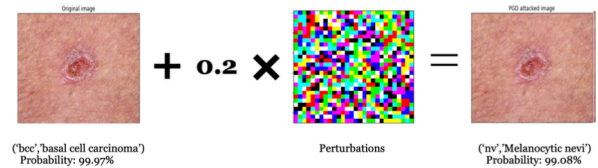

Figure 2: PGD attack

In figure 2, on the left-hand side, the same dermatoscopic image which was used to demonstrate the FGSM attack can be observed. This image is classified by the model to be a 'basal cell carcinoma' with a probability of 99.97%. The perturbation is generated using a PGD attack and is multiplied with a multiplication factor and added to the original image. The image generated from the PGD attack is now classified as 'melanocytic nevi' with a probability of 99.08%. This attack has fooled the model into misclassifying the adversarial example with a confidence of 99.08%.

### V. ADVERSARIAL TRAINING

There are numerous strategies that can be used to protect models from adversarial attacks. Adversarial training is one of the most effective and widely used techniques for deep learning models. The goal of adversarial training is to increase the inherent robustness [8] of the model. This is done by generating a variety of attacks against the model proactively and preparing the system to recognise potential adversarial attacks. This process is similar to building up the 'immunity' or 'tolerance' of the model to these attacks, the model can perform better against adversarial examples as a result, and the rate of misclassification of such images decreases drastically.

### VI. METHODOLOGY

We conducted our experiments on a CNN model trained to categorize dermatoscopic images of skin lesions in order to comprehend the effects of these adversarial attacks on machine learning models and the usefulness of the defense techniques previously described in the context of medical image classification systems.

We first evaluate the performance of our vanilla CNN model (unexposed to adversarial examples in the training phase) against adversarial examples generated by both FGSM and PGD. Following that we conduct adversarial training in a variety of ways, and the performance of the model trained using each training technique is compared across adversarial examples of both FGSM and PGD techniques. The process is described in greater detail below.

FGSM attack:

The FGSM attack is carried out in accordance with equation 1. We have taken the value of ε to be 0.2, which is the factor with which the created perturbations are multiplied before adding them to the original images. We conducted this attack on our test dataset to obtain a dataset of adversarial examples generated using FGSM. The performance of our vanilla model is then evaluated against these attacked images.

PGD attack:

We now conduct the PGD attack, the working of which is described in equation 2. As seen in the equation, 'δ' represents the perturbation generated, which is multiplied with the factor 'α' in order to make the attacked image look indistinguishable from the original.

For our implementation, we have taken 'α' to be 0.2, and ε to be 0.3. The number of iterations for the purpose of our attack is 50. This attack is carried out on the original test data, thus generating a dataset of adversarial examples obtained using PGD. Our vanilla model's performance against this dataset is then evaluated.

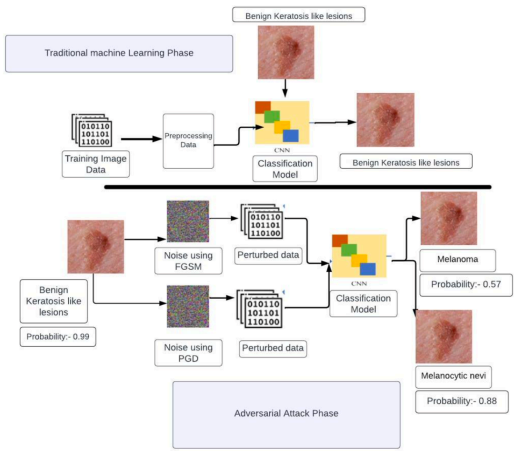

Figure 3: Adversarial attacks

Adversarial Training:

In order to demonstrate the increase in robustness observed when employing adversarial training, we have carried out this defensive technique on our model. The standard procedure to train a deep neural network can be seen in equation 3, while equation 4 illustrates the approach we have taken to training our model in the process of conducting adversarial training.

$$minE(x, y)\varepsilon\ Dataset\ [L(x, y; \theta)] \quad (3)$$

$$minE(x, y)\varepsilon\ Dataset[\ \underset{||\delta||\ p\ \leq \varepsilon}{max\ L}\ (x + \delta, y; \theta)] \quad (4)$$

To determine the best method to increase our model's robustness against the attacks seen earlier, we have adopted the following process. We employ adversarial training in three different approaches: training our model on adversarial examples that have been crafted using FGSM, training our model on adversarial examples created using PGD, and on an even combination of adversarial examples that have been generated using FGSM and PGD respectively.

We add adversarial instances produced from a portion of the training dataset for each strategy to the original training dataset. The hostile examples produced from the randomly chosen fraction are combined with the original data to create the training data.

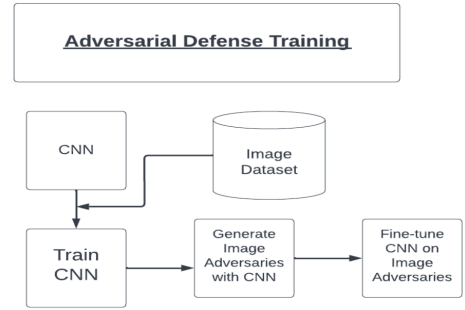

Figure 4: Adversarial training

To determine the optimal concentration of these adversarial examples, we train our model on three different concentrations and compare the results. We generate adversarial examples for 20%, 40%, 60%, and 80% of the training data respectively, comparing the results in each instance. The model weights are randomized before each instance of training to ensure an accurate evaluation. As a result, adversarial training is conducted using FGSM, PGD, and an equal number of adversarial instances created by FGSM and PGD. In the final instance, adversarial instances are produced using FGSM on half of the randomly chosen data that will be attacked and PGD on the other half. After then, these instances are blended to create some of the training data.

VII. RESULTS AND ANALYSIS

The following provides a detailed explanation of the outcomes of the experiments conducted on our paper and how these outcomes were analyzed.

Table 1. Comparison of performance of the vanilla CNN model on the original and adversarially attacked dataset.

| Accuracy Of Vanilla Model | Original Test Data (x_test) | Test data attacked using FGSM | Test data attacked using PGD |
|---|---|---|---|
| | 70.19% | 57.21% | 37.69% |

Table 1 illustrates the accuracy of the vanilla model on the test data (x_test), FGSM images, and PGD images respectively. As seen in the table, the accuracy of the model drops greatly from 70.19% to 57.21%, and 37.69% on FGSM and PGD respectively. Hence, it is observed that the PGD attack leads to a greater misclassification rate as compared to FGSM.

The reasons for this are stated as follows:

- In the PGD attack, the attack image is computed using a number of iterations, as compared to the FGSM attack where only iteration is utilized.
- This shifts the gradients with respect to the input image at a greater rate as compared to FGSM which leads to a greater loss maximization.
- A greater loss maximization indicates a greater misclassification rate.

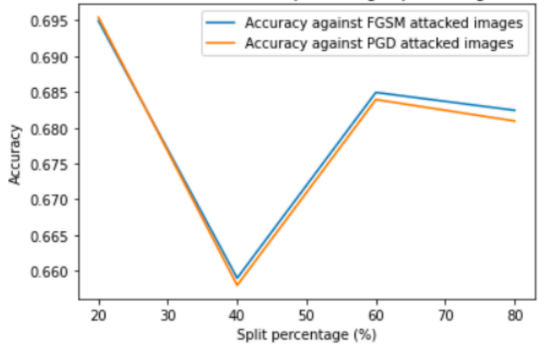

Figure 5: FGSM adversarial training

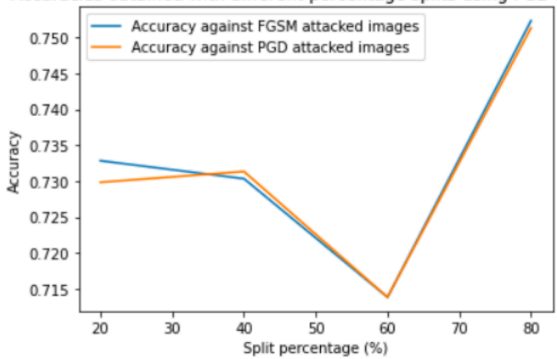

Figure 6: PGD adversarial training

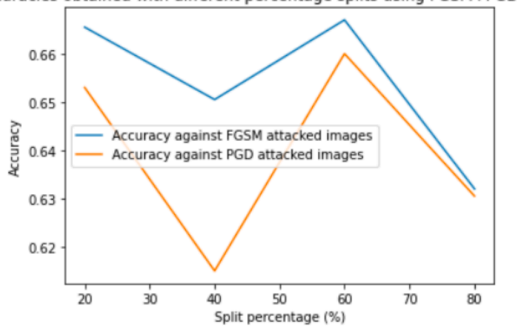

Figure 7: Ensemble adversarial training

The experimental analysis on the various adversarial training procedures that was done for this paper is represented graphically in Figures 5, 6, and 7. As seen in the figures, the training was most effective at different concentrations of adversarial examples for different techniques. The model-wise comparison has been represented in the table below, where the best performing adversarial training split for each approach has been considered.

Table 2. Comparison of performance of adversarial training approaches with their respective optimal training split.

| Model | Original Test Data | Test Data Attacked Using FGSM | Test Data Attacked Using PGD | Optimal Adversarial Training Split |
|---|---|---|---|---|
| Trained on FGSM examples | 69.64% | 69.49% | 69.54% | 20% |
| Trained on PGD examples | 75.33% | 75.23% | 75.13% | 80% |
| Trained on FGSM & PGD examples | 67.34% | 66.70% | 66.00% | 60% |

Table 2 illustrates the accuracy obtained using each defense method on its most optimal and best performing split. It can be observed that PGD training performs better than FGSM training. As described earlier, the PGD attack is a more effective adversarial attack. It thus follows that using PGD training enables the model to learn the underlying concepts related to classification better, making it more robust to adversarial examples. Hence, better accuracy as a result of PGD training can be observed.

It's noteworthy to observe that neither of the other two strategies used in this paper are as good at defending the model as the training method that combines the adversarial cases produced by PGD and FGSM. One possible cause could be the confusion created by mixing the attacked feature vectors generated with both PGD and FGSM, each of which represent different learning behavior, leading to this deficit in performance.

VIII. CONCLUSION

In this paper we analyzed popular techniques to perform adversarial attacks in the context of the medical field. We employed the use of a CNN trained to classify skin lesions from dermatoscopic images in order to demonstrate the vulnerability of medical image classification systems against gradient-based adversarial attacks. The usefulness of adversarial training as a method of defending against such attacks was then examined. Thus we are able to gain an understanding of the usefulness of adversarial training against such white-box attacks to be employed in order to increase the resilience of systems for classifying medical images.

REFERENCES

1. Ching, T., Himmelstein, D. S., Beaulieu-Jones, B. K., Kalinin, A. A., Do, B. T., Way, G. P., ... & Greene, C. S. (2018). Opportunities and obstacles for deep learning in biology and medicine. *Journal of The Royal Society Interface*, *15*(141), 20170387.

2. Goodfellow, I. J., Shlens, J., & Szegedy, C. (2014). Explaining and harnessing adversarial examples. arXiv preprint arXiv:1412.6572.

3. P. Tschandl, C. Rosendahl, and H. Kittler, "The HAM10000 dataset, a large collection of multi-source dermatoscopic images of common pigmented skin lesions", Scientific Data, vol. 5, pp. 180161, 2018.

4. Madry, A., Makelov, A., Schmidt, L., Tsipras, D., & Vladu, A. (2017). Towards deep learning models resistant to adversarial attacks. arXiv preprint arXiv:1706.06083.

5. Shafahi, A., Najibi, M., Ghiasi, M. A., Xu, Z., Dickerson, J., Studer, C., ... & Goldstein, T. (2019). Adversarial training for free!. Advances in Neural Information Processing Systems, 32.

6. Dong, Y., Liao, F., Pang, T., Su, H., Zhu, J., Hu, X., & Li, J. (2018). Boosting adversarial attacks with momentum. In Proceedings of the IEEE conference on computer vision and pattern recognition (pp. 9185-9193).

7. Finlayson, S.G., Chung, H.W., Kohane, I.S., Beam, A.L.: Adversarial attacks against medical deep learning systems. arXiv preprint arXiv:1804.05296 (2018)

8. Chen, H.Y., Liang, J.H., Chang, S.C., Pan, J.Y., Chen, Y.T., Wei, W., Juan, D.C.: Improving adversarial robustness via guided complement entropy. In: Proceedings of the IEEE International Conference on Computer Vision. pp. 4881–4889 (2019).

9. Ren, K., Zheng, T., Qin, Z., & Liu, X. (2020). Adversarial attacks and defenses in deep learning. *Engineering*, *6*(3), 346-360.

10. Liu, X., Faes, L., Kale, A. U., Wagner, S. K., Fu, D. J., Bruynseels, A., ... & Denniston, A. K. (2019). A comparison of deep learning performance against healthcare professionals in detecting diseases from medical imaging: a systematic review and meta-analysis. *The lancet digital health*, *1*(6), e271-e297.

11. Kermany, D. S., Goldbaum, M., Cai, W., Valentim, C. C., Liang, H., Baxter, S. L., ... & Zhang, K. (2018). Identifying medical diagnoses and treatable diseases by image-based deep learning. *Cell*, *172*(5), 1122-1131.

12. Qin, Y., Zheng, H., Huang, X., Yang, J., & Zhu, Y. M. (2019). Pulmonary nodule segmentation with CT sample synthesis using adversarial networks. *Medical physics*, *46*(3), 1218-1229.

13. Moosavi-Dezfooli SM, Fawzi A, Fawzi O, Frossard P. Universal adversarial perturbations. In: Proceedings of the 2017 IEEE Conference on Computer Vision and Pattern Recognition; 2017 Jul 21–26; Honolulu, HI, USA; 2017. p. 1765–73.

14. Madry A, Makelov A, Schmidt L, Tsipras D, Vladu A. Towards deep learning models resistant to adversarial attacks. 2017. arXiv: 1706.06083